\documentclass[letterpaper]{article} 
\usepackage{aaai2026}  
\nocopyright
\usepackage{times}  
\usepackage{helvet}  
\usepackage{courier}  
\usepackage[hyphens]{url}  
\usepackage{graphicx} 
\usepackage{natbib}  
\usepackage{caption} 
\usepackage{algorithm}
\usepackage{algorithmic}
\usepackage{verbatim}
\usepackage{amsmath}
\usepackage{amssymb}
\usepackage{tabularx}
\usepackage{makecell}
\usepackage{subcaption}
\usepackage{graphicx}
\usepackage{multirow}
\usepackage{booktabs}

\makeatletter
\title{Dynamic Participation in Federated Learning: Benchmarks and a Knowledge Pool Plugin
\begingroup
\footnote{Accepted at the $1^{st}$ Workshop on Federated Learning for Critical Applications (FLCA), in conjunction with AAAI26}
\addtocounter{footnote}{-1}
\endgroup}
\makeatother

 \author {
     Ming-Lun Lee\textsuperscript{\rm 1},
     Fu-Shiang Yang\textsuperscript{\rm 1},
     Cheng-Kuan Lin\textsuperscript{\rm 1},
     \\
     Yan-Ann Chen\textsuperscript{\rm 2},
     Chih-Yu Lin\textsuperscript{\rm 3},
    Yu-Chee Tseng\textsuperscript{\rm 1}
 }
 \affiliations {
     \textsuperscript{\rm 1}Department of Computer Science,
National Yang Ming Chiao Tung University, Hsinchu, Taiwan\\
     \textsuperscript{\rm 2}Department of Computer Science and Engineering, Yuan Ze University, Taoyuan, Taiwan\\
     \textsuperscript{\rm 3}Department of Computer Science and Engineering, National Taiwan Ocean University, Keelung, Taiwan\\
 }

\usepackage{bibentry}

\begin{document}

\maketitle

\begin{abstract}
Federated learning (FL) enables clients to collaboratively train a shared model in a distributed manner, setting it apart from traditional deep learning paradigms. 
However, most existing FL research assumes consistent client participation, overlooking the practical scenario of dynamic participation (DPFL), where clients may intermittently join or leave during training. 
Moreover, no existing benchmarking framework systematically supports the study of DPFL-specific challenges.
In this work, we present the first open-source framework explicitly designed for benchmarking FL models under dynamic client participation. 
Our framework provides configurable data distributions, participation patterns, and evaluation metrics tailored to DPFL scenarios. 
Using this platform, we benchmark four major categories of widely adopted FL models and uncover substantial performance degradation under dynamic participation.
To address these challenges, we further propose Knowledge-Pool Federated Learning (KPFL), a generic plugin that maintains a shared knowledge pool across both active and idle clients. 
KPFL leverages dual-age and data-bias weighting, combined with generative knowledge distillation, to mitigate instability and prevent knowledge loss. 
Extensive experiments demonstrate the significant impact of dynamic participation on FL performance and the effectiveness of KPFL in improving model robustness and generalization.
\end{abstract}

\begin{links}
    \link{Code}{https://github.com/NYCU-PAIR-Labs/DPFL}
\end{links}



\begin{figure}[h]
    \centering
    \includegraphics[width=0.85\linewidth]{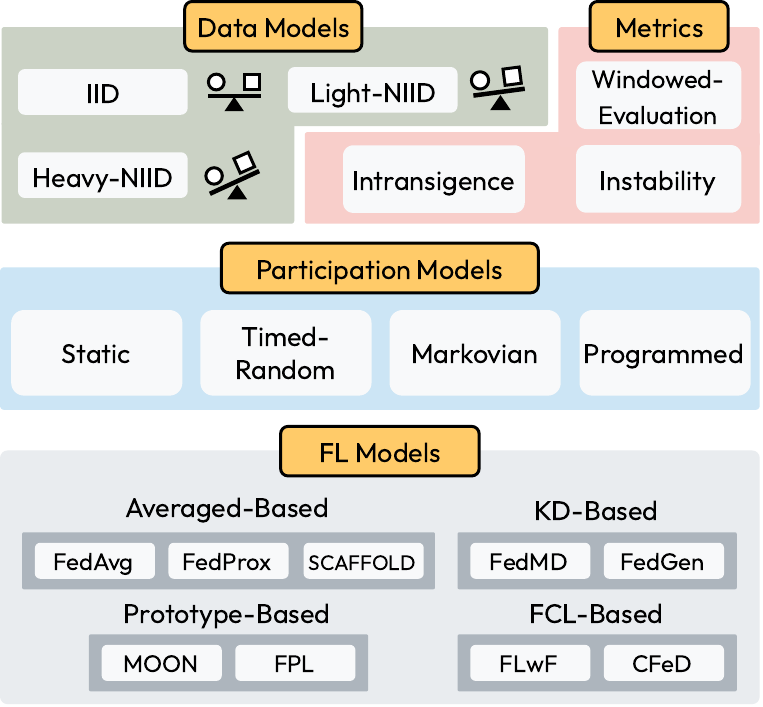}
    \caption{The proposed DPFL benchmarking framework.}
    \label{fig:framework}
\end{figure}

\section{Introduction}
Federated learning (FL) \cite{fedavg} enables collaborative training of a server model across multiple clients without sharing private data.
However, most FL solutions assume that clients participate consistently throughout the training process.
In mobile or unstable environments, clients may dynamically join, leave, and even rejoin training.
Such dynamic participation (DP) is common among battery-powered, solar-operated edge devices, as well as participants in social or vehicular networks~\cite{FL_VIoT}.

To address this realistic yet underexplored setting, we investigate dynamic participation in federated learning (DPFL).
DP introduces several key challenges:
First, it significantly degrades training performance due to inconsistent data availability and highly variable client updates, which can slow convergence~\cite{gu2021fast, ruan2021towards}, increase instability~\cite{fedrem}, and result in catastrophic forgetting~\cite{1999catastrophic, goodfellow2015, 2017overcoming, FCL}.
Second, participation patterns are inherently unpredictable, as some clients drop out temporarily or permanently, and some join only in late stages.
Third, the non-IID nature of client data further exacerbates these issues~\cite{NonIID}.

Dynamic client behaviors have been partially addressed in previous works.
Flexible device participation is considered in \cite{ruan2021towards}, which studies how inactive devices and mid-training arrivals or departures affect convergence and extends FedAvg~\cite{fedavg} with simple weighting.
MIFA \cite{gu2021fast} mitigates client unavailability by caching gradients but applies static aggregation weights.

Federated Continual Learning (FCL) \cite{flwf, cfed, FCL} focuses on preserving and adapting knowledge across evolving tasks. FCL methods handle consistent feature space expansion when new tasks are introduced.
In contrast, DPFL addresses the unstable feature space and prevents knowledge collapse caused by fragmented updates from intermittent clients.
Learning dynamics have also been explored in FedAvg and Asynchronous Federated Learning (AFL) \cite{afl_survey, afl_1, afl_2}. FedAvg uses random client sampling to scale with large client pools, while AFL accommodates varying client availability to improve aggregation efficiency. 
In addition, recent effort \cite{flp} prototypes FL as a digital learning platform.
However, these works do not explicitly model the participation patterns that characterize DPFL.
Although several platforms exist, such as TFF~\cite{tff}, FedML~\cite{fedml}, 
Flower~\cite{flower}, and PySyft~\cite{pysyft}, they lack support for DP scenarios.
To our knowledge, there is currently no benchmark capable of evaluating the unique challenges posed by DPFL.

In this work, we fill this gap by introducing the first open-source DPFL platform that explicitly models client dynamics (Fig.~\ref{fig:framework}). The platform provides (i) controllable data models (e.g., IID, light- and heavy-NIID), (ii) realistic and probabilistic DP models, and (iii) a set of DPFL-specific evaluation metrics. Our benchmark suite systematically evaluates the impact of DPFL across nine state-of-the-art FL models spanning four widely-used categories. 

Building on insights from this benchmark, we propose knowledge-pool FL (KPFL), a generic plugin that maintains historical model states for both active and idle clients. By applying age-aware and data-bias–aware weighting with generative distillation, KPFL substantially mitigates instability and knowledge loss. We demonstrate successful integration of KPFL into all nine evaluated models, achieving substantial improvements in robustness under various DP scenarios.

\section{Review: Benchmarking Targets}
\label{sec:review}

We benchmark 9 FL methods in 4 groups. (i) {\em average-based} methods: FedAvg~\cite{fedavg}, FedProx~\cite{fedprox}, and SCAFFOLD~\cite{scaffold}, (ii) {\em knowledge distillation (KD)-based} methods: FedMD~\cite{fedmd} and FedGen~\cite{fedgen}, (iii) {\em prototype-based} methods: MOON~\cite{moon} and FPL~\cite{fpl}, (iv) {\em federated continual learning (FCL)-based} methods: FLwF~\cite{flwf} and CFeD~\cite{cfed}.

\section{Design of the DPFL Framework}
\label{sec:framework}

\subsection{Modeling Data Heterogeneity}  \label{sec:data_model}

To introduce data heterogeneity, we modify the label distributions across clients and classes following the Dirichlet distribution \cite{NonIID}. 
For each class $k$, we sample $p_k \sim \text{Dir}_N(\alpha)$, where $N$ is the number of clients, $p_k$ is a vector of length $N$, and $\alpha$ is the concentration parameter. Client $j$ receives a proportion $p_{k,j}$ of the class-$k$ data, $j = 1,2,\ldots, N$. 
Based on $\alpha$, we categorize data heterogeneity into three levels: 
(i) {\em IID}: $\alpha = 100$; the distributions are approximately uniform.
(ii) {\em Light-NIID}: $\alpha = 1.0$; the distributions are moderately non-IID.
(iii) {\em Heavy-NIID}: $\alpha = 0.1$; the distributions are highly imbalanced.

\subsection{Modeling Participation Dynamics}

Let $C = \{c_0, c_1, \ldots, c_{N-1}\}$ denote the complete set of clients, where $c_i$ represents the $i$-th client, $i = 0,1, \ldots, N-1$. Let $C_t \subseteq C$ be the subset of clients participating in training round $t$. 
Our platform supports several models.

\textbf{Timed-Random.} 
Each client $c_i$ participates in round $t$ with a time-varying probability $p_i(t)$, modeling geographically distributed, solar-powered clients.

\textbf{Markovian.} 
Each client alternates between active and inactive states based on a Markov transition matrix 
$\mathcal{P} = \left[\begin{smallmatrix}
p_{0\to0} & p_{0\to1} \\ 
p_{1\to0} & p_{1\to1}
\end{smallmatrix}\right]$
, modeling energy-saving Discontinuous Reception (DRX) \cite{drx_1, drx_2, drx_3} in LTE/5G.

\textbf{Programmed.} 
Custom participation patterns can be imported, allowing replay of real-world client behaviors.

\subsection{Performance Metrics}  \label{sec:metrics}
While current FL benchmarks primarily focus on final-round accuracy, we define several cross-round quantitative metrics to better understand the impact of DPFL.

\textbf{Windowed Evaluation ($\uparrow$):} 
Let $\psi_i$ denote a performance metric of the global model measured in the $i$-th round. The windowed evaluation metric, ${W\!E}_t$, is computed as a statistical property of $\psi_i$ (e.g., mean, variance; mean by default) over a window of $\omega$ consecutive rounds.

\textbf{Intransigence to DP ($\downarrow$):}
This metric quantifies the average performance gap between dynamic and static participation, reflecting a model’s robustness to client dynamics.

\begin{equation}
I\!D\!P = \frac{1}{T_{f}} \textstyle \sum_{i=1}^{T_{f}} 
(\psi^*_i - \psi_i)
\end{equation}

\textbf{Instability due to DP ($\downarrow$):}
This metric assesses the stability of a given performance index $\psi_i$ within a window $\textstyle i \in [T_{s_1}: T_{s_2}]$ by calculating the average absolute deviation between the actual and its regressed values $\{\tilde{\psi}_i\}$.
\begin{equation}
I\!D_{T_{s_1}, T_{s_2}} = 
\textstyle \left(
\sum_{i=T_{s_1}+1}^{T_{s_2}} 
| \tilde{\psi}_i - \psi_i | \right)
/
(T_{s_2} - T_{s_1})
\end{equation}

\section{Knowledge-Pool Federated Learning}
\begin{figure} 
    \includegraphics[width=\linewidth]{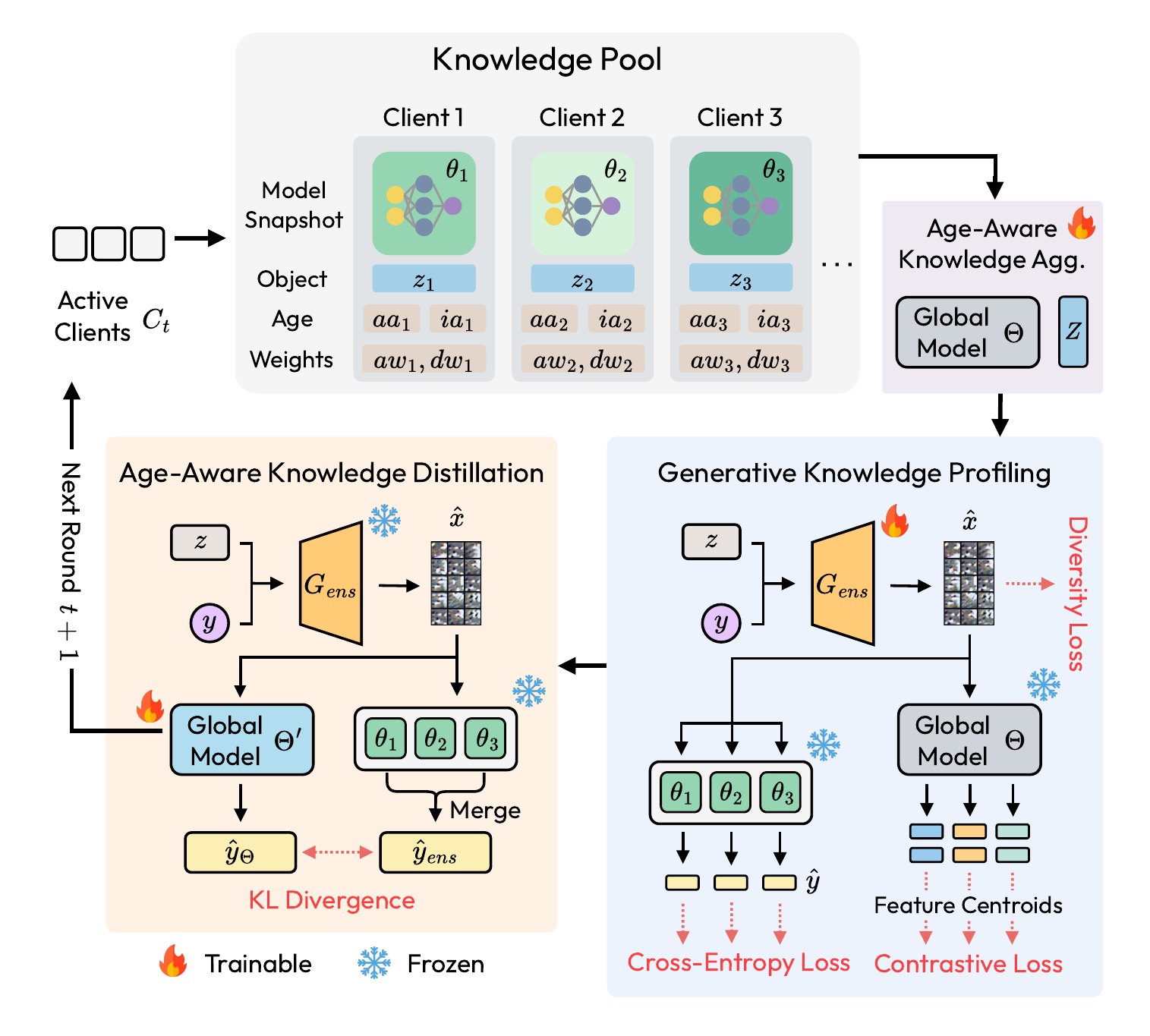}
    \caption{The Knowledge-Pool Federated Learning framework.}
    \label{fig:kpfl-framework}
\end{figure}

KPFL is designed as a model-agnostic plug-in to mitigate DPFL challenges while remaining fully compatible with existing FL models.
Fig.~\ref{fig:kpfl-framework} shows the architecture of KPFL.

\subsection{Knowledge Pool}

The server maintains a knowledge pool $\mathcal{KP}$, which captures the evolving knowledge of both active and idle clients to support adaptive and robust global model training.
For each client $c_i$, the server maintains the following information:

\begin{table}[h]
\small
\centering
\setlength{\tabcolsep}{1.5pt} 
\begin{tabular}{|c|p{6.5cm}|}
\hline
$\theta_i$ & the most recent local model update of $c_i$ \\ 
\hline
$\tau_i$ & the timestamp of $c_i$'s most recent state transition (active/idle) \\
\hline
$n_{i,j}$ & the number of data items of class $j$ held by $c_i$ \\
\hline
$z_i$ & the local auxiliary object associated with $\theta_i$ \\
\hline
${aa}_i$, ${ia}_i$ & active age and idle age indicators \\
\hline
${aw}_i$, ${dw}_i$ & age weight and data bias weight \\
\hline
\end{tabular}
\label{tab:information}
\end{table}

We propose a {\em dual-age weighting} mechanism to address adaptiveness and forgetfulness.
For each active client $c_i$, we define its active age as $aa_i = t - \tau_i$, where $t$ is the current time (or round). A smaller $aa_i$ implies greater freshness.
Conversely, for each idle client $c_i$, we define its idle age as $ia_i = t - \tau_i$, where a larger idle age indicates more severe staleness.
The age weight of client $c_i$ at time $t$ is given by:
\begin{equation}
    aw_i = 
    \begin{cases}
    e^{\lambda_{aa} \cdot aa_{i}} & 
    \text{if $c_i$ is active} 
    \\
    e^{\lambda_{ia} \cdot ia_{i}} & 
    \text{if $c_i$ is idle}
    \end{cases}
\end{equation}
where $\lambda_{aa}$ and $\lambda_{ia}$ are the age decay coefficients.

To further balance statistical heterogeneity, each client $c_i$ is assigned a data bias weight $dw_i = \sum_{j=1}^{K} {n_{i, j}}/N_{j}$,
where $N_j = \sum_{c_i} n_{i, j}$ is the total number of class $j$ samples. This design allows clients with larger representative datasets to contribute proportionally more during aggregation.

\subsection{Age-Aware Knowledge Aggregation}
Via the age and data bias weights, we aggregate the first version of global model $\Theta$ and a global auxiliary object $\mathcal{Z}$. 
\begin{align}
(\Theta, \mathcal{Z}) &=  \textstyle \sum_{c_i} 
    {W}_i \cdot (\theta_i, z_i)
\end{align}
${W}_i =      aw_{i} \cdot \varepsilon_{i}
    + dw_{i}$ represents an aggregated weight of $c_i$ using a weighting factor 
$\varepsilon_{i} = \sum_j n_{i, j} /\sum_{i, j} n_{i, j}$.

\subsection{Age-Aware Generative Knowledge Distillation}

We introduce an age-aware generative knowledge profiling method that mitigates knowledge degradation under DPFL.
Specifically, we train a conditional generator $G_{ens}$ to synthesize collective knowledge from both active and idle clients in the knowledge pool $\mathcal{KP}$.
Given a latent vector $z \sim \mathcal{N}(0, 1)$ and a class label $y$ sampled from $p(y) = N_y / \sum_{i,j} n_{i,j}$, $y = 1,2,\ldots,K$, the generator produces class-conditioned samples $\hat{x} = G_{ens}(z, y)$.

To ensure that $\hat{x}$ reflects meaningful and diverse knowledge, we optimize $G_{ens}$ using three loss functions. 
The first is an age-weighted cross-entropy loss:
\begin{equation}
    \mathcal{L}_{ce} = \textstyle \mathbb{E}_{\hat{x}} 
            \left[ \, \sum_{c_i} {W}_i^{y} 
           \cdot
           CE(\theta_i (\hat{x}), y)  \, \right]
\end{equation}
where weight ${W}_i^{y} = aw_{i} +\textstyle \frac{n_{i, y}}{N_{y}}$ accounts for model staleness of $c_i$ on data class $y$:

Second, to ensure feature-level discriminability, we apply contrastive learning to supervise the feature space via $\Theta$:
\begin{equation}
    \mathcal{L}_{ctr} = - \log \frac{ exp(d(f(\hat x), cen(y))/ \tau)}{\sum_{y'} exp(d(f(\hat x), cen(y'))/ \tau)}
\end{equation}
where $f(\cdot)$ is the feature extractor of $\Theta$, $cen(y)$ is the semantic center of class $y$, $d(\cdot)$ is a similarity function (e.g., cosine similarity), $y'$ iterates over all classes, and $\tau$ is a temperature parameter.

To prevent model collapse and promote diversity in $G_{ens}$, 
we include a data diversity loss term $\mathcal{L}_{div}$, following the design in \cite{fedftg, fedgen}.

To summarize, the overall objective is to optimize $G_{ens}$ with the loss
$\mathcal{L} = \gamma_{ce} \mathcal{L}_{ce} + \gamma_{ctr} \mathcal{L}_{ctr} +  \gamma_{div} \mathcal{L}_{div}$,
where $\gamma_{ce}$, $\gamma_{ctr}$, and $\gamma_{div}$ are weighting factors.

After training the generator, we fine-tune $\Theta$ by distilling age-weighted knowledge from the ensemble generated by $G_{ens}$ to obtain an improved global model $\Theta '$:
\begin{align}
    \hat{y}_{ens} &= \textstyle \sum_{c_i} {W}_i^{y} \cdot \hat{y}_i
   \\
    \hspace*{-0.8em}  \mathcal{L}_{ens} \!&=\! 
   \mathbb{E}_{\hat{x}} 
                    \left[ \, K\!L(\sigma (\hat{y}_{ens}) || \sigma (\hat{y}_{\Theta})) \, \right]
\end{align}

\begin{table*}[t]
\centering
\fontsize{9pt}{10pt}\selectfont

\begin{tabular}{cc|c|ccccccccc|r}
\Xhline{1.5pt} 
& Metric               & Type  & FedAvg & FedProx & SCAF. & FedMD & FedGen & MOON  & FPL   & FLwF  & CFeD  & AVG   \\ \hline
\multirow{8}{*}{w/o KPFL}  & \multirow{3}{*}{$W\!E (\uparrow)$}  & Stat. & 63.02  & 64.99   & 63.20 & 52.75 & 73.23  & 63.07 & 58.19 & 65.33 & 57.98 & 62.42 \\
 &                    & T-R   & 46.01  & 47.83   & 50.03 & 34.50 & 56.05  & 46.40 & 46.15 & 46.34 & 51.74 & 47.23 \\
 &                    & M     & 50.92  & 53.30   & 51.32 & 33.59 & 54.75  & 52.44 & 51.28 & 53.22 & 53.05 & 50.43 \\ \cline{2-13}
& \multirow{2}{*}{$I\!D\!P (\downarrow)$} & T-R   & 20.87  & 20.57   & 17.82 & 15.80 & 20.65  & 21.06 & 17.95 & 21.19 & 11.40 & 18.59 \\
  &                   & M     & 19.12  & 18.25   & 15.80 & 17.19 & 21.95  & 19.11 & 15.72 & 19.02 & 11.62 & 17.53 \\ \cline{2-13}
& \multirow{3}{*}{$I\!D (\downarrow)$}  & Stat. & 0.49   & 0.46    & 0.38  & 0.54  & 0.29   & 0.52  & 0.69  & 0.43  & 0.29  & 0.45  \\
 &                    & T-R   & 1.58   & 1.74    & 1.44  & 1.24  & 1.72   & 1.72  & 1.52  & 1.68  & 1.49  & 1.57  \\
  &                   & M     & 1.11   & 1.22    & 1.19  & 1.02  & 1.24   & 1.31  & 1.03  & 1.18  & 1.10  & 1.16  \\ \hline
                                                
\multirow{8}{*}{w/ KPFL}  & \multirow{3}{*}{$W\!E (\uparrow)$}  & Stat. & \underline{66.79}  & 66.06   & \underline{66.75} & 53.51 & \textbf{72.17}  & 65.59 & 58.70 & 65.93 & 65.46 & 64.55 \\
 &                    & T-R   & 60.57  & \underline{62.49}   & 61.75 & 49.04 & \textbf{65.00}  & 61.31 & 54.06 & 59.85 & \underline{63.12} & 59.69 \\
  &                   & M     & 58.10  & \underline{59.88}   & 57.79 & 44.15 & \textbf{62.66}  & \underline{60.50} & 53.26 & 59.05 & 58.25 & 57.07 \\ \cline{2-13}
& \multirow{2}{*}{$I\!D\!P (\downarrow)$} & T-R   & 4.78   &\underline{ 4.34}    & 5.02  & \textbf{1.13}  & 9.86   & 5.35  & 5.31  & 6.29  & \underline{4.38}  & 5.16  \\
  &                   & M     & 8.17   & \underline{6.82}    & 8.63  & \textbf{5.39}  & 11.18  & 7.91  & \underline{6.30}  & 8.03  & 8.80  & 7.91  \\ \cline{2-13}
& \multirow{2}{*}{$I\!D (\downarrow)$}  & T-R   & 0.52   & \underline{0.43}    & 0.48  & 0.47  & 0.45   & \textbf{0.36}  & 0.48  & \textbf{0.36}  & 0.47  & 0.45  \\
  &                   & M     & 0.50   & 0.45    & 0.52  & \textbf{0.36}  & 0.47   & \underline{0.41}  & 0.51  & \underline{0.40}  & 0.47  & 0.45   \\
\Xhline{1.5pt} 
\end{tabular}
\caption{Benchmarking result on Heavy-NIID Office-Caltech.}
\label{tab:final_result_part}
\end{table*}

\section{Experiment Results}
\subsection{Benchmarking Setups}

We developed an open-source benchmark platform that supports all DPFL scenarios defined in Sec.~\ref{sec:framework} and benchmarked all 9 FL models reviewed in Sec.~\ref{sec:review}.
We conducted experiments on image classification dataset \textit{Office-Caltech} \cite{fpl}.
The dataset is partitioned following the settings in Sec.~\ref{sec:framework}. Due to space limits, we present the key results focusing on the most challenging setting, Heavy-NIID Office-Caltech, in the main paper, which is most affected by DP due to high data heterogeneity and limited client overlap. 
Full experimental results on other datasets 
and settings are provided on the benchmarking platform.

We adopted the widely used ResNet-10 \cite{resnet} with a 512-dimensional feature vector for training.
Experiments were conducted with 10--50 clients, using a batch size of 128 and running for $T_f = 100$ rounds, each containing 5 local epochs.
SGD optimizer was used with a learning rate of 0.01, weight decay of $1 \times 10^{-5}$, and momentum of 0.9.

Configurations of the DP patterns are as follows:
(i) Timed-random: each client has a joining probability of $p_i(t) = 0.5$.
(ii) Markovian: transition probabilities are set as $[p_{0 \to 0}, p_{0 \to 1}] = [0.8, 0.2]$ and $[p_{1 \to 0}, p_{1 \to 1}] = [0.2, 0.8]$.

\subsection{Performance Benchmarking under DPFL}

The following results are averaged over 5 runs with different random seeds.

\textbf{Final $W\!E$ with $\omega = 5$:}
As shown in Tab.~\ref{tab:final_result_part}, while most FL methods perform reliably under static setting, the introduction of DP severely degrades model performance. Under Timed-Random and Markovian, all methods failed to retain existing knowledge as clients drop out and rejoin, resulting in accuracy losses of around 15\% and 12\%, respectively.

\textbf{Intransigence to DP ($I\!D\!P$):} All evaluated methods yield positive $I\!D\!P$ scores, showing that DP consistently impairs convergence against static baselines. For example, FedAvg yields a score of 20.87 on Timed-Random and 19.12 on Markovian, revealing large convergence gaps caused by DP.

\textbf{Instability due to DP ($I\!D$):} 
Static participation maintains stable learning with an average $I\!D$ score of 0.45. In contrast, highly dynamic participation, particularly Timed-Random, leads to the highest instability of 1.57.
Robustness methods also struggled, FedProx and FedGen reach $I\!D$ of 1.7 on Timed-Random, and MOON records 1.3 on Markovian.

As shown, DP significantly degrades all existing FL methods across metrics. No existing method demonstrates reliable robustness, highlighting a critical gap in handling DP.

\subsection{KPFL Plugin: Effectiveness and Impact}
\textbf{Fully compatible to existing FL models.}
We successfully integrated KPFL into all 9 FL models reviewed in Sec.~\ref{sec:review}.
Tab.~\ref{tab:final_result_part} below presents the results of KPFL-enhanced models. Across all DP types, plugging in KPFL consistently improves final accuracy $W\!E$, achieving more than 12\% gain under Timed-Random.
KPFL performs most outstandingly in $I\!D\!P$, 
narrowing the convergence gap from 18.59$\rightarrow$5.16 under Timed-Random and from 17.53$\rightarrow$7.91 under Markovian.
KPFL also restores stability, reducing the $I\!D$ scores to the level of static participation.

\noindent\textbf{Scaling KPFL with Larger Client Pools.}
Tab.~\ref{tab:scale} reports the scalability of KPFL on top of FedProx. As the client pool increases from 20 to 50, KPFL consistently improves performance across all metrics. While larger client pools exhibit greater stability, KPFL further reinforces learning robustness, demonstrating its scalability to larger client scales. 

These results highlight both the limitations of existing methods and the effectiveness of the proposed $\mathcal{KP}$ design in managing varying client availability.

\begin{table}[t]
\centering
\setlength{\tabcolsep}{4.3pt} 
{\fontsize{9pt}{10pt}\selectfont
\begin{tabular}{cccccccc}
\Xhline{1.5pt}
                     & \multicolumn{3}{c}{$W\!E (\uparrow)$}                 & \multicolumn{2}{c}{$I\!D\!P (\downarrow)$}       & \multicolumn{2}{c}{$I\!D (\downarrow)$} \\ \hline
\multicolumn{1}{c|}{}        & Stat.          & T-R            & \multicolumn{1}{c|}{M}              & T-R           & \multicolumn{1}{c|}{M}             & T-R           & M             \\ \hline
\multicolumn{8}{c}{number of clients $=20$}                                                                                                                                                                  \\ \hline
\multicolumn{1}{c|}{FedProx} & 59.48          & 51.50          & \multicolumn{1}{c|}{45.50}          & 13.64         & \multicolumn{1}{c|}{11.63}         & 1.15          & 1.48          \\
\multicolumn{1}{c|}{+KPFL}   & \textbf{60.58} & \textbf{57.41} & \multicolumn{1}{c|}{\textbf{54.13}} & \textbf{4.23} & \multicolumn{1}{c|}{\textbf{9.20}} & \textbf{0.39} & \textbf{0.49} \\ \hline
\multicolumn{8}{c}{number of clients $=50$}                                                                                                                                                                  \\ \hline
\multicolumn{1}{c|}{FedProx} & 51.41          & 45.20          & \multicolumn{1}{c|}{41.50}          & 9.17          & \multicolumn{1}{c|}{8.93}          & 1.12          & 0.99          \\
\multicolumn{1}{c|}{+KPFL}   & \textbf{53.01} & \textbf{50.30} & \multicolumn{1}{c|}{\textbf{48.26}} & \textbf{3.59} & \multicolumn{1}{c|}{\textbf{7.31}} & \textbf{0.29} & \textbf{0.42}   \\

\Xhline{1.5pt}
\end{tabular}
}
\caption{Performance gains of KPFL-enhanced FedProx under Heavy-NIID Office-Caltech.
}
\label{tab:scale}
\end{table}

\begin{table}
\centering
{\fontsize{9pt}{10pt}\selectfont
\setlength{\tabcolsep}{4.0pt} 
\begin{tabular}{cccccccc}
\Xhline{1.5pt}

                    &               & \multicolumn{3}{c}{FedProx + } &   \multicolumn{3}{c}{MOON + } \\ 
         \cmidrule(lr){3-5}
         \cmidrule(lr){6-8}
&         &  \multirow{2}{*}{MIFA}     & \multicolumn{2}{c}{ KPFL} & \multirow{2}{*}{MIFA}  &\multicolumn{2}{c}{ KPFL} 

\\            
         \cmidrule(lr){4-5}
         \cmidrule(lr){7-8}
 & &   & $\Theta$ & $\Theta'$ &   & $\Theta$ & $\Theta'$    \\
\hline
\multirow{3}{*}{$W\!E(\uparrow)$} 
& Stat.    &   64.04   & 65.79 & \textbf{66.06} &       63.57 & 65.29 & \textbf{65.59} \\
& T-R      &   60.55   & 60.28 & \textbf{62.49} &       59.15 & 60.98 & \textbf{61.31} \\ 
& M        &   \textbf{61.03}   & 57.89 & 59.88 &       59.71 & 57.55 & \textbf{60.50} \\ 
\hline
\multirow{2}{*}{$I\!D(\downarrow)$}
& T-R      &   0.46 & \textbf{0.42}  & 0.43 &      0.51 & 0.47  & \textbf{0.36}   \\ 
& M        &   0.40 & \textbf{0.38}  & 0.45 &      0.42 & 0.46  & \textbf{0.41}   \\ 

\Xhline{1.5pt}
\end{tabular}
}
\caption{
Ablation study on $\Theta$ and $\Theta '$ (Heavy-NIID Office-Caltech). MIFA is included as an alternative aging approach.}
\label{tab:ablation}
\end{table}

\subsection{Ablation Study}
In KPFL, there is an initial global model $\Theta$ and an improved $\Theta '$.
Tab.~\ref{tab:ablation} shows our ablation study results. 
MIFA \cite{gu2021fast} is included as a baseline, as it also stores the latest gradients for idle clients but applies static weights. 
In contrast, KPFL employs a dual-age strategy, which explicitly takes idle and active clients' states into account.
The results show that $\Theta '$ consistently outperforms $\Theta$, validating the benefit of our two-stage design.
Additionally, comparison with Tab.~\ref{tab:final_result_part} highlights the individual contributions and overall efficacy of KPFL.
Furthermore, $\Theta '$ outperforms MIFA in most cases, showing substantial improvements in learning robustness, which demonstrates the effectiveness of our design.

\section{Conclusions}
\label{sec:con}

This paper presents several benchmarking scenarios to systematically investigate the challenges of DPFL. Our study demonstrates that DP significantly impacts the learning efficiency of a wide range of existing FL models across diverse datasets. Furthermore, the proposed KPFL plugin effectively mitigates the effects of participation dynamics across various FL model types.

\section*{Acknowledgements}
This work was supported by grants of National Science and Technology Council (NSTC 114-2221-E-A49-11), and the NVIDIA Academic Grant Program 2025. 

\bibliography{main}

\end{document}